\documentclass{article}
\usepackage{arxiv}

\usepackage[utf8]{inputenc} 
\usepackage[T1]{fontenc}    
\usepackage{hyperref}       
\usepackage{url}            
\usepackage{booktabs}       
\usepackage{amsfonts}       
\usepackage{nicefrac}       
\usepackage{microtype}      
\usepackage{lipsum}
\usepackage[numbers, sort, comma, square]{natbib}
\usepackage{fancyhdr}       
\usepackage{graphicx}       
\graphicspath{{media/}}     
\usepackage{amsmath}
\usepackage{amssymb}
\usepackage{tikz}
\usepackage{float} 
\usepackage{stfloats}
\usetikzlibrary{arrows.meta, positioning, fit, calc}

\definecolor{latB}{HTML}{0072B2}
\definecolor{latB}{HTML}{0072B2}\definecolor{latG}{HTML}{999999}
\title{LexLattice: Multilingual Extractive Summarization via Neural Cellular Automata on Document Hierarchies
}

\author{
  Sujay Uday Rittikar \\
  Applied Computer Science \\
  The University of Winnipeg \\
  Winnipeg, MB \\
  \texttt{rittikar-s@webmail.uwinnipeg.ca} \\
  \And
  Sheela Ramanna \\
  Applied Computer Science \\
  The University of Winnipeg \\
  Winnipeg, MB \\
  \texttt{s.ramanna@uwinnipeg.ca} \\
}

\begin{document}
\maketitle

\begin{abstract}
Faithfulness is a central concern in legal text summarization, which motivates extractive approaches that select verbatim content traceable to its source. Such methods typically rank paragraphs or other structural units in isolation, yet give little attention to \emph{consolidating} evidence that is distributed across, and shares salience between, distant parts of a document. We introduce LexLattice, an extractive summarizer that reifies a legal act's hierarchy as a two-dimensional semantic lattice and consolidates over it with a masked 2D neural cellular automata before selection. LexLattice attains state-of-the
art ROUGE across all 24 languages of EUR-Lex-Sum in both multilingual and cross-lingual settings, surpassing instruction-tuned baselines with billions of parameters, despite concentrating all trainable capacity in a 1.8M parameter
consolidator over a frozen multilingual encoder. A consolidator trained only on high-resource languages further transfers to unseen languages with near-lossless retention (0.99), indicating that the model operates on language-agnostic semantic geometry rather than surface form. Our results position explicit consolidation over document structure as a compact and traceable alternative to scale for multilingual legal summarization.
\end{abstract}


\section{Introduction}

The European Union (EU) enacts its legislation in 24 official languages, and a single legal act often runs to tens of thousands of tokens of dense, formulaic prose that binds not only legal professionals but the citizens and businesses subject to it. Automatic summarization is therefore a natural instrument for access to law. It is an important tool to turn lengthy legislative instruments into concise, faithful accounts of what they require to present. However, this is a demanding setting: Documents run well beyond standard encoder limits, and a system must work uniformly across languages that differ sharply in available resources. Faithfulness is the sharper of these pressures \citep{feijo2023improving}, and it is what makes extractive approaches especially attractive here: a summary assembled from verbatim provisions can be traced back to the exact article it came from, and it must not invent legal content that the act does not contain. Yet the standard neural extractive recipe encodes the document and then scores each sentence in reading order, as one long flat sequence \citep{nallapati2017summarunner, liu-lapata-2019-text}, discarding structure that legal drafting marks out explicitly. An EU act is not a stream of sentences but a hierarchy of recitals, articles, paragraphs, and annexes, and a passage's position within it says a great deal about whether the passage belongs in a summary: the first paragraph of an operative article typically states an obligation, a recital supplies background reasoning, an annex lists technical detail. A model must infer these roles from wording alone, even though the document already declares them \citep{cohan-etal-2018-discourse, 10.1145/3545176, ruan-etal-2022-histruct}.

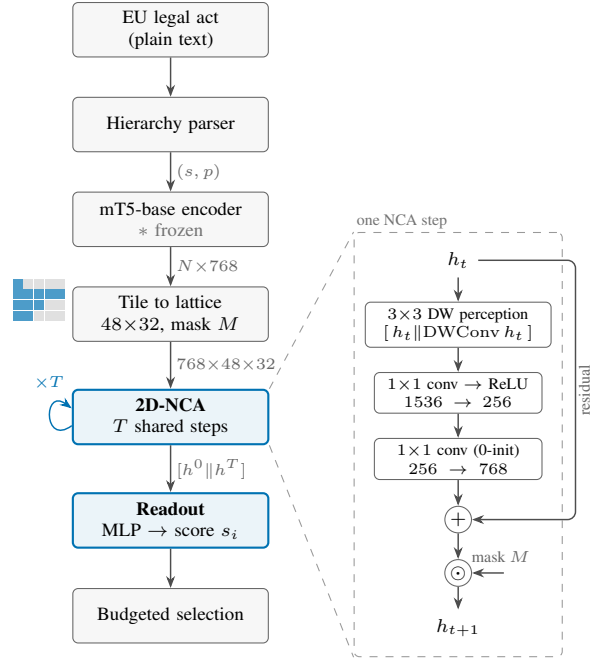
\begin{figure}[t]
	\centering
	\begin{tikzpicture}[
		blk/.style={draw=black!60, fill=black!3, rounded corners=2pt, align=center,
			font=\scriptsize, inner sep=2.6pt, text width=2.40cm,
			minimum height=0.72cm},
		acc/.style={blk, draw=latB, fill=latB!8, line width=0.8pt},
		ins/.style={draw=black!60, fill=white, rounded corners=2pt, align=center,
			font=\tiny, inner sep=2.2pt, text width=2.06cm},
		op/.style={draw=black!60, fill=white, circle, inner sep=0.7pt, font=\scriptsize},
		arr/.style={-{Stealth[length=1.6mm]}, black!70, line width=0.55pt},
		alab/.style={font=\tiny, text=black!60, inner sep=1pt, anchor=west},
		dsh/.style={black!40, dashed, line width=0.4pt}]
		
		\node[blk] (doc) at (0,0) {EU legal act (plain text)};
		\node[blk] (par) [below=0.52cm of doc] {Hierarchy parser};
		\node[blk] (enc) [below=0.52cm of par] {mT5-base encoder\\\textcolor{black!55}{$\ast$ frozen}};
		\node[blk] (til) [below=0.52cm of enc] {Tile to lattice\\$48{\times}32$, mask $M$};
		\node[acc] (nca) [below=0.62cm of til] {\textbf{2D-NCA}\\$T$ shared steps};
		\node[acc] (rd)  [below=0.62cm of nca] {\textbf{Readout}\\MLP $\to$ score $s_i$};
		\node[blk] (sel) [below=0.52cm of rd] {Budgeted selection};
		
		\draw[arr] (doc) -- (par);
		\draw[arr] (par) -- node[alab, right=1pt] {$(s,p)$} (enc);
		\draw[arr] (enc) -- node[alab, right=1pt] {$N{\times}768$} (til);
		\draw[arr] (til) -- node[alab, right=1pt] {$768{\times}48{\times}32$} (nca);
		\draw[arr] (nca) -- node[alab, right=1pt] {$[h^0 \Vert h^T]$} (rd);
		\draw[arr] (rd)  -- (sel);
		
		\draw[arr, latB] (nca.185) .. controls +(-0.38,-0.28) and +(-0.38,0.28) .. (nca.175);
		\node[font=\tiny, text=latB, anchor=south] at ($(nca.west)+(-0.30,0.24)$) {$\times T$};
		
		\begin{scope}[shift={($(til.west)+(-0.78,0.34)$)}]
			\foreach \r/\n in {0/1, 1/5, 2/3, 3/2}{
				\foreach \c in {0,...,4}{
					\pgfmathtruncatemacro{\occ}{\c<\n ? 1 : 0}
					\ifnum\occ=1
					\fill[latB!70] (\c*0.14,-\r*0.14) rectangle ++(0.12,0.12);
					\else
					\fill[black!12] (\c*0.14,-\r*0.14) rectangle ++(0.12,0.12);
					\fi
				}
			}
		\end{scope}
		
		\node[font=\scriptsize] (h0) at (3.80,-3.05) {$h_t$};
		\node[ins, text width=2.30cm] (perc) [below=0.32cm of h0]
		{$3{\times}3$ DW perception\\$[\,h_t \Vert \mathrm{DWConv}\,h_t\,]$};
		\node[ins] (w1) [below=0.32cm of perc] {$1{\times}1$ conv $\to$ ReLU\\$1536 \to 256$};
		\node[ins] (w2) [below=0.32cm of w1] {$1{\times}1$ conv (0-init)\\$256 \to 768$};
		\node[op]  (plus) [below=0.34cm of w2] {$+$};
		\node[op]  (mask) [below=0.34cm of plus] {$\odot$};
		\node[font=\scriptsize] (h1) [below=0.30cm of mask] {$h_{t+1}$};
		
		\draw[arr] (h0) -- (perc);
		\draw[arr] (perc) -- (w1);
		\draw[arr] (w1) -- (w2);
		\draw[arr] (w2) -- (plus);
		\draw[arr] (plus) -- (mask);
		\draw[arr] (mask) -- (h1);
		\coordinate (railx) at (5.35,0);
		\coordinate (labx)  at (5.47,0);
		\path ($(h0.east)!0.5!(plus.east)$) coordinate (rmid);
		\draw[arr, rounded corners=3pt]
		(h0.east) -- (railx |- h0.east) -- (railx |- plus.east) -- (plus.east);
		\node[font=\tiny, text=black!60, rotate=90, anchor=center, inner sep=0pt]
		at (labx |- rmid) {residual};
		\draw[arr] ($(mask.east)+(0.42,0)$) -- (mask.east);
		\node[font=\tiny, text=black!60, anchor=south, inner sep=1pt]
		at ($(mask.east)+(0.38,0.10)$) {mask $M$};
		
		\node[draw=black!40, dashed, rounded corners=2pt, inner sep=4pt,
		fit=(h0)(perc)(w1)(w2)(plus)(mask)(h1)] (inset) {};
		\node[font=\tiny, text=black!55, anchor=south west, inner sep=1pt]
		at (inset.north west) {one NCA step};
		
		\draw[dsh] (nca.north east) -- (inset.north west);
		\draw[dsh] (nca.south east) -- (inset.south west);
	\end{tikzpicture}
	\caption{The LexLattice architecture. A rule-based parser (Section \ref{subsec:parsing}) recovers the act's hierarchy; a frozen mT5-base encoder embeds each paragraph, and embeddings are tiled onto the semantic lattice (cf.\ Fig.~\ref{fig:lattice}). The masked 2D-NCA applies one shared local update $T{=}8$ times (right: one step), and a small readout scores each
		un-tiled paragraph for budgeted selection. Blue blocks are the only trainable components ($\approx$1.8M parameters); the mask $M$ keeps padding out of every
		update.}
	\label{fig:model}
\end{figure}

In this work, we propose \textbf{LexLattice}, an extractive summarizer that accounts for document structure by reifying it as a two-dimensional \textbf{semantic lattice}: rows index the sections of an act, columns index paragraph position within a section, and each cell carries the frozen multilingual encoder state \citep{xue-etal-2021-mt5} of the corresponding text unit. Over this lattice we run a masked two-dimensional Neural Cellular Automata (NCA) \citep{mordvintsev2020growing}, a parameter-light convolutional update rule applied iteratively, so that salience evidence propagates locally \textit{along} sections (between adjacent paragraphs) and \textit{across} them (between structurally parallel positions) before any selection decision is made. Our hypothesis is that document structure, once made geometric, allows a small recurrent local model perform the global evidence consolidation that sequence models otherwise, obtain through long-range self-attention and far heavier parameterization. We evaluate on EUR-Lex-Sum across all 24 languages in multilingual and all-pairs cross-lingual settings, comparing a capacity-matched 1D-NCA control, the 2D LexLattice, and its reinforcement-tuned LexLattice + RLOO. With only 1.8M trainable parameters over a frozen mT5-base encoder, LexLattice + RLOO achieves the best ROUGE among all systems, including recent LLM and SLM baselines, while remaining competitive on BERTScore-F1.

\section{Related Works}
\label{sec:rw}

The EUR-Lex-Sum benchmark \citep{aumiller-etal-2022-eur} provides reference baselines in all languages from a zero-shot multilingual LexRank extractor \citep{Erkan2004}, and includes a cross-lingual English-to-Spanish track evaluated with translate-then-summarize pipelines over OPUS-MT \citep{tiedemann-thottingal-2020-opus} and a Longformer Encoder-Decoder \citep{beltagy2020longformerlongdocumenttransformer}. \citet{sie-etal-2024-summarizing} propose a multi-step extractive-then-abstractive pipeline for long regulatory documents, coupling a LexLM extractor to LongT5 \citep{guo-etal-2022-longt5}, Pegasus \citep{pmlr-v119-zhang20ae}, or a QLoRA-tuned Llama-3 decoder \citep{grattafiori2024llama3herdmodels}. Their two-stage design first extracts salient content and then rewrites it abstractively, improving fluency and coherence over the extracted intermediate summary. The extractive stage that precedes it remains essential for keeping the generator's input faithful to the source, a concern that becomes central in the legal domain. While the above mentioned systems focus on the summarization pipeline, \citet{t-y-s-s-etal-2024-lexsumm} instead emphasize the model: they introduce LexT5, a sequence-to-sequence model pre-trained, fine-tuned, and probed on legal-domain knowledge, and benchmark it against Longformer and PRIMERA \citep{xiao-etal-2022-primera} variants. Complementing these, \citet{bendahman-etal-2025-hallucinations} presented an analysis of entity-level faithfulness on the same corpus, illustrating distinction between expert-introduced abstractive entities and spurious hallucinations. Recently, the trend of Small Language Models (SLMs) has encouraged various works on text summarization \citep{chheda-etal-2025-extract, divide_summarize_slm}. A study by \citet{slms_legislative_sum} benchmarked the small instruction-tuned language models against the large language models (LLMs) on six languages of EUR-Lex-Sum across an extensive list of models. Their results indicate that instruction tuning alone outperforms task fine-tuning under the LLM-as-judge evaluations.

\section{Dataset}
\label{sec:dataset}

\begin{table*}[t]
	\centering
	\small
	\setlength{\tabcolsep}{5pt}
	\caption{EUR-Lex-Sum statistics per language after structural pre-processing. Validation and test splits contain 187 and 188 (act, summary) pairs for every language. Word, section, and paragraph counts are means over the training split. \emph{Tier} marks the cross-lingual transfer split: we train on the 12 high-resource (H) languages and evaluate zero-shot on the 12 low-resource (L) languages. The bottom row gives the total training documents and macro-averages over languages. $^{\dagger}$Irish: 16 training documents, and the longest source documents in the corpus.}	
	\begin{tabular}{@{}llrrrrr@{}}
		\toprule
		\textbf{Language} & \textbf{Tier} & \textbf{Training} & \textbf{Document} & \textbf{Summary} & \textbf{Sections} & \textbf{Paragraphs} \\
		& & \textbf{documents} & \textbf{words} & \textbf{words} & \textbf{per document} & \textbf{per document} \\
		\midrule
		Bulgarian    & L &    957 & 13,219 &   861 & 38.8 & 150.9 \\
		Croatian     & L &    903 & 11,043 &   765 & 41.8 & 172.1 \\
		Czech        & H &    984 & 12,072 &   757 & 28.3 & 113.8 \\
		Danish       & H &  1,002 & 11,677 &   768 & 28.7 & 110.7 \\
		Dutch        & H &  1,001 & 12,979 &   887 & 28.4 & 112.0 \\
		English      & H &  1,129 & 12,079 &   820 & 29.2 & 115.7 \\
		Estonian     & L &    957 & 10,460 &   625 & 25.1 & 110.4 \\
		Finnish      & H &    991 &  9,662 &   629 & 26.7 & 104.1 \\
		French       & H &  1,130 & 13,665 &   993 & 28.7 & 108.0 \\
		German       & H &  1,115 & 10,897 &   799 & 26.2 &  72.2 \\
		Greek        & H &    993 & 13,301 &   904 & 27.3 & 113.7 \\
		Hungarian    & L &    961 & 11,862 &   757 & 27.2 &  89.9 \\
		Irish$^{\dagger}$ & L & 16 & 32,852 & 1,020 & 30.9 & 168.0 \\
		Italian      & H &  1,028 & 13,174 &   915 & 27.8 & 107.9 \\
		Latvian      & L &    959 & 11,443 &   714 & 25.0 & 124.7 \\
		Lithuanian   & L &    960 & 11,499 &   707 & 24.6 & 104.6 \\
		Maltese      & L &    940 & 14,745 &   917 & 28.6 & 119.1 \\
		Polish       & L &    978 & 11,702 &   786 & 26.7 & 105.8 \\
		Portuguese   & H &  1,001 & 13,070 &   945 & 28.3 & 112.2 \\
		Romanian     & L &    978 & 13,485 &   875 & 39.7 & 104.5 \\
		Slovak       & L &    950 & 11,699 &   764 & 30.1 & 125.2 \\
		Slovenian    & L &    957 & 11,485 &   755 & 30.0 & 121.6 \\
		Spanish      & H &  1,112 & 13,257 &   975 & 26.1 &  98.7 \\
		Swedish      & H &    987 & 11,123 &   763 & 29.1 & 119.6 \\
		\midrule
		Total / mean & & 22,989 & 13,019 & 821 & 29.3 & 116.1 \\
		\bottomrule
	\end{tabular}
	\label{tab:dataset}
\end{table*}

We use EUR-Lex-Sum \citep{aumiller-etal-2022-eur}, a manually curated
multilingual and cross-lingual dataset of legal acts and their summaries, drawn from the EUR-Lex law platform and covering all 24 official EU languages. We use the released train/validation/test partitions in full, without language or document sampling. Table~\ref{tab:dataset} reports per-language statistics after pre-processing. Each record consists of a CELEX identifier \citep{celex2006}, the full text of the act, and the reference summary, all as plain newline-delimited text. The release carries no structural markup, thus, the document structure must be recovered from the text itself.

\subsection{Rule-based hierarchy parsing}
\label{subsec:parsing}

We recover each act's hierarchy with a deterministic, rule-based parser built on EU legal drafting conventions, using a manually verified keyword table for all languages covering the head-words for \textbf{article}, \textbf{chapter}, \textbf{section}, \textbf{title}, and \textbf{annex}. The article patterns accommodate both \textit{word-first layouts} ("Article 1", "Artigo 1.º") and \textit{number-first layouts} ("1 artikla", "1. cikk"); recitals are detected by the pan-European "(N)" line convention and numbered paragraphs by leading "N." markers. A line is considered as a potential header only if it is shorter than 60 characters, guarding against body text that merely contains a keyword. The parser emits a list of sections, each holding an ordered list of paragraphs. Only four section kinds become structural units downstream: \textbf{preamble}, the \textbf{recitals block}, \textbf{articles}, and \textbf{annexes}, while chapter/section/title headings act purely as paragraph-flow separators. Sections with no paragraphs are discarded.

\paragraph{Text units.} The atomic unit for selection is the \textit{paragraph}, addressed by its coordinate (section index, paragraph index). For encoding only, paragraphs longer than 256 multilingual T5 (mT5) \citep{xue-etal-2021-mt5} tokens are split into consecutive 256-token chunks whose embeddings are later mean-pooled back to a single paragraph vector. Chunking therefore stays internal to encoding: the model still selects, and is supervised on, whole paragraphs rather than chunks, so the unit of prediction matches the unit of the oracle target.

\section{Preliminaries}
\label{sec:prelim}

\subsection{The Semantic Lattice}

\begin{figure}[t]
	\centering
	\definecolor{latPre}{HTML}{CC79A7}
	\definecolor{latRec}{HTML}{0072B2}
	\definecolor{latArt}{HTML}{009E73}
	\definecolor{latAnx}{HTML}{D55E00}
	\definecolor{latPad}{HTML}{EDEDED}
	\begin{tikzpicture}[x=1cm,y=1cm]
		\fill[latPre,draw=white,line width=0.5pt] (0.00,-0.40) rectangle (0.40,0.00); \fill[latPad,draw=white,line width=0.5pt] (0.40,-0.40) rectangle (0.80,0.00); \fill[latPad,draw=white,line width=0.5pt] (0.80,-0.40) rectangle (1.20,0.00); \fill[latPad,draw=white,line width=0.5pt] (1.20,-0.40) rectangle (1.60,0.00); \fill[latPad,draw=white,line width=0.5pt] (1.60,-0.40) rectangle (2.00,0.00); \fill[latPad,draw=white,line width=0.5pt] (2.00,-0.40) rectangle (2.40,0.00); \fill[latPad,draw=white,line width=0.5pt] (2.40,-0.40) rectangle (2.80,0.00); \fill[latPad,draw=white,line width=0.5pt] (2.80,-0.40) rectangle (3.20,0.00); \fill[latPad,draw=white,line width=0.5pt] (3.20,-0.40) rectangle (3.60,0.00); \fill[latPad,draw=white,line width=0.5pt] (3.60,-0.40) rectangle (4.00,0.00); \fill[latPad,draw=white,line width=0.5pt] (4.00,-0.40) rectangle (4.40,0.00); \fill[latPad,draw=white,line width=0.5pt] (4.40,-0.40) rectangle (4.80,0.00);
		\node[anchor=east,font=\scriptsize,inner sep=1.5pt] at (-0.06,-0.20) {Preamble};
		\fill[latRec,draw=white,line width=0.5pt] (0.00,-0.80) rectangle (0.40,-0.40); \fill[latRec,draw=white,line width=0.5pt] (0.40,-0.80) rectangle (0.80,-0.40); \fill[latRec,draw=white,line width=0.5pt] (0.80,-0.80) rectangle (1.20,-0.40); \fill[latRec,draw=white,line width=0.5pt] (1.20,-0.80) rectangle (1.60,-0.40); \fill[latRec,draw=white,line width=0.5pt] (1.60,-0.80) rectangle (2.00,-0.40); \fill[latRec,draw=white,line width=0.5pt] (2.00,-0.80) rectangle (2.40,-0.40); \fill[latRec,draw=white,line width=0.5pt] (2.40,-0.80) rectangle (2.80,-0.40); \fill[latRec,draw=white,line width=0.5pt] (2.80,-0.80) rectangle (3.20,-0.40); \fill[latRec,draw=white,line width=0.5pt] (3.20,-0.80) rectangle (3.60,-0.40); \fill[latRec,draw=white,line width=0.5pt] (3.60,-0.80) rectangle (4.00,-0.40); \fill[latRec,draw=white,line width=0.5pt] (4.00,-0.80) rectangle (4.40,-0.40); \fill[latRec,draw=white,line width=0.5pt] (4.40,-0.80) rectangle (4.80,-0.40);
		\node[anchor=east,font=\scriptsize,inner sep=1.5pt] at (-0.06,-0.60) {Recitals};
		\fill[latArt,draw=white,line width=0.5pt] (0.00,-1.20) rectangle (0.40,-0.80); \fill[latPad,draw=white,line width=0.5pt] (0.40,-1.20) rectangle (0.80,-0.80); \fill[latPad,draw=white,line width=0.5pt] (0.80,-1.20) rectangle (1.20,-0.80); \fill[latPad,draw=white,line width=0.5pt] (1.20,-1.20) rectangle (1.60,-0.80); \fill[latPad,draw=white,line width=0.5pt] (1.60,-1.20) rectangle (2.00,-0.80); \fill[latPad,draw=white,line width=0.5pt] (2.00,-1.20) rectangle (2.40,-0.80); \fill[latPad,draw=white,line width=0.5pt] (2.40,-1.20) rectangle (2.80,-0.80); \fill[latPad,draw=white,line width=0.5pt] (2.80,-1.20) rectangle (3.20,-0.80); \fill[latPad,draw=white,line width=0.5pt] (3.20,-1.20) rectangle (3.60,-0.80); \fill[latPad,draw=white,line width=0.5pt] (3.60,-1.20) rectangle (4.00,-0.80); \fill[latPad,draw=white,line width=0.5pt] (4.00,-1.20) rectangle (4.40,-0.80); \fill[latPad,draw=white,line width=0.5pt] (4.40,-1.20) rectangle (4.80,-0.80);
		\node[anchor=east,font=\scriptsize,inner sep=1.5pt] at (-0.06,-1.00) {Article ~1};
		\fill[latArt,draw=white,line width=0.5pt] (0.00,-1.60) rectangle (0.40,-1.20); \fill[latPad,draw=white,line width=0.5pt] (0.40,-1.60) rectangle (0.80,-1.20); \fill[latPad,draw=white,line width=0.5pt] (0.80,-1.60) rectangle (1.20,-1.20); \fill[latPad,draw=white,line width=0.5pt] (1.20,-1.60) rectangle (1.60,-1.20); \fill[latPad,draw=white,line width=0.5pt] (1.60,-1.60) rectangle (2.00,-1.20); \fill[latPad,draw=white,line width=0.5pt] (2.00,-1.60) rectangle (2.40,-1.20); \fill[latPad,draw=white,line width=0.5pt] (2.40,-1.60) rectangle (2.80,-1.20); \fill[latPad,draw=white,line width=0.5pt] (2.80,-1.60) rectangle (3.20,-1.20); \fill[latPad,draw=white,line width=0.5pt] (3.20,-1.60) rectangle (3.60,-1.20); \fill[latPad,draw=white,line width=0.5pt] (3.60,-1.60) rectangle (4.00,-1.20); \fill[latPad,draw=white,line width=0.5pt] (4.00,-1.60) rectangle (4.40,-1.20); \fill[latPad,draw=white,line width=0.5pt] (4.40,-1.60) rectangle (4.80,-1.20);
		\node[anchor=east,font=\scriptsize,inner sep=1.5pt] at (-0.06,-1.40) {Article ~2};
		\fill[latArt,draw=white,line width=0.5pt] (0.00,-2.00) rectangle (0.40,-1.60); \fill[latArt,draw=white,line width=0.5pt] (0.40,-2.00) rectangle (0.80,-1.60); \fill[latArt,draw=white,line width=0.5pt] (0.80,-2.00) rectangle (1.20,-1.60); \fill[latPad,draw=white,line width=0.5pt] (1.20,-2.00) rectangle (1.60,-1.60); \fill[latPad,draw=white,line width=0.5pt] (1.60,-2.00) rectangle (2.00,-1.60); \fill[latPad,draw=white,line width=0.5pt] (2.00,-2.00) rectangle (2.40,-1.60); \fill[latPad,draw=white,line width=0.5pt] (2.40,-2.00) rectangle (2.80,-1.60); \fill[latPad,draw=white,line width=0.5pt] (2.80,-2.00) rectangle (3.20,-1.60); \fill[latPad,draw=white,line width=0.5pt] (3.20,-2.00) rectangle (3.60,-1.60); \fill[latPad,draw=white,line width=0.5pt] (3.60,-2.00) rectangle (4.00,-1.60); \fill[latPad,draw=white,line width=0.5pt] (4.00,-2.00) rectangle (4.40,-1.60); \fill[latPad,draw=white,line width=0.5pt] (4.40,-2.00) rectangle (4.80,-1.60);
		\node[anchor=east,font=\scriptsize,inner sep=1.5pt] at (-0.06,-1.80) {Article ~3};
		\fill[latArt,draw=white,line width=0.5pt] (0.00,-2.40) rectangle (0.40,-2.00); \fill[latArt,draw=white,line width=0.5pt] (0.40,-2.40) rectangle (0.80,-2.00); \fill[latArt,draw=white,line width=0.5pt] (0.80,-2.40) rectangle (1.20,-2.00); \fill[latPad,draw=white,line width=0.5pt] (1.20,-2.40) rectangle (1.60,-2.00); \fill[latPad,draw=white,line width=0.5pt] (1.60,-2.40) rectangle (2.00,-2.00); \fill[latPad,draw=white,line width=0.5pt] (2.00,-2.40) rectangle (2.40,-2.00); \fill[latPad,draw=white,line width=0.5pt] (2.40,-2.40) rectangle (2.80,-2.00); \fill[latPad,draw=white,line width=0.5pt] (2.80,-2.40) rectangle (3.20,-2.00); \fill[latPad,draw=white,line width=0.5pt] (3.20,-2.40) rectangle (3.60,-2.00); \fill[latPad,draw=white,line width=0.5pt] (3.60,-2.40) rectangle (4.00,-2.00); \fill[latPad,draw=white,line width=0.5pt] (4.00,-2.40) rectangle (4.40,-2.00); \fill[latPad,draw=white,line width=0.5pt] (4.40,-2.40) rectangle (4.80,-2.00);
		\node[anchor=east,font=\scriptsize,inner sep=1.5pt] at (-0.06,-2.20) {Article ~4};
		\fill[latArt,draw=white,line width=0.5pt] (0.00,-2.80) rectangle (0.40,-2.40); \fill[latArt,draw=white,line width=0.5pt] (0.40,-2.80) rectangle (0.80,-2.40); \fill[latArt,draw=white,line width=0.5pt] (0.80,-2.80) rectangle (1.20,-2.40); \fill[latPad,draw=white,line width=0.5pt] (1.20,-2.80) rectangle (1.60,-2.40); \fill[latPad,draw=white,line width=0.5pt] (1.60,-2.80) rectangle (2.00,-2.40); \fill[latPad,draw=white,line width=0.5pt] (2.00,-2.80) rectangle (2.40,-2.40); \fill[latPad,draw=white,line width=0.5pt] (2.40,-2.80) rectangle (2.80,-2.40); \fill[latPad,draw=white,line width=0.5pt] (2.80,-2.80) rectangle (3.20,-2.40); \fill[latPad,draw=white,line width=0.5pt] (3.20,-2.80) rectangle (3.60,-2.40); \fill[latPad,draw=white,line width=0.5pt] (3.60,-2.80) rectangle (4.00,-2.40); \fill[latPad,draw=white,line width=0.5pt] (4.00,-2.80) rectangle (4.40,-2.40); \fill[latPad,draw=white,line width=0.5pt] (4.40,-2.80) rectangle (4.80,-2.40);
		\node[anchor=east,font=\scriptsize,inner sep=1.5pt] at (-0.06,-2.60) {Article ~5};
		\fill[latArt,draw=white,line width=0.5pt] (0.00,-3.20) rectangle (0.40,-2.80); \fill[latPad,draw=white,line width=0.5pt] (0.40,-3.20) rectangle (0.80,-2.80); \fill[latPad,draw=white,line width=0.5pt] (0.80,-3.20) rectangle (1.20,-2.80); \fill[latPad,draw=white,line width=0.5pt] (1.20,-3.20) rectangle (1.60,-2.80); \fill[latPad,draw=white,line width=0.5pt] (1.60,-3.20) rectangle (2.00,-2.80); \fill[latPad,draw=white,line width=0.5pt] (2.00,-3.20) rectangle (2.40,-2.80); \fill[latPad,draw=white,line width=0.5pt] (2.40,-3.20) rectangle (2.80,-2.80); \fill[latPad,draw=white,line width=0.5pt] (2.80,-3.20) rectangle (3.20,-2.80); \fill[latPad,draw=white,line width=0.5pt] (3.20,-3.20) rectangle (3.60,-2.80); \fill[latPad,draw=white,line width=0.5pt] (3.60,-3.20) rectangle (4.00,-2.80); \fill[latPad,draw=white,line width=0.5pt] (4.00,-3.20) rectangle (4.40,-2.80); \fill[latPad,draw=white,line width=0.5pt] (4.40,-3.20) rectangle (4.80,-2.80);
		\node[anchor=east,font=\scriptsize,inner sep=1.5pt] at (-0.06,-3.00) {Article ~6};
		\fill[latArt,draw=white,line width=0.5pt] (0.00,-3.60) rectangle (0.40,-3.20); \fill[latArt,draw=white,line width=0.5pt] (0.40,-3.60) rectangle (0.80,-3.20); \fill[latArt,draw=white,line width=0.5pt] (0.80,-3.60) rectangle (1.20,-3.20); \fill[latArt,draw=white,line width=0.5pt] (1.20,-3.60) rectangle (1.60,-3.20); \fill[latPad,draw=white,line width=0.5pt] (1.60,-3.60) rectangle (2.00,-3.20); \fill[latPad,draw=white,line width=0.5pt] (2.00,-3.60) rectangle (2.40,-3.20); \fill[latPad,draw=white,line width=0.5pt] (2.40,-3.60) rectangle (2.80,-3.20); \fill[latPad,draw=white,line width=0.5pt] (2.80,-3.60) rectangle (3.20,-3.20); \fill[latPad,draw=white,line width=0.5pt] (3.20,-3.60) rectangle (3.60,-3.20); \fill[latPad,draw=white,line width=0.5pt] (3.60,-3.60) rectangle (4.00,-3.20); \fill[latPad,draw=white,line width=0.5pt] (4.00,-3.60) rectangle (4.40,-3.20); \fill[latPad,draw=white,line width=0.5pt] (4.40,-3.60) rectangle (4.80,-3.20);
		\node[anchor=east,font=\scriptsize,inner sep=1.5pt] at (-0.06,-3.40) {Article ~7};
		\fill[latArt,draw=white,line width=0.5pt] (0.00,-4.00) rectangle (0.40,-3.60); \fill[latArt,draw=white,line width=0.5pt] (0.40,-4.00) rectangle (0.80,-3.60); \fill[latArt,draw=white,line width=0.5pt] (0.80,-4.00) rectangle (1.20,-3.60); \fill[latArt,draw=white,line width=0.5pt] (1.20,-4.00) rectangle (1.60,-3.60); \fill[latPad,draw=white,line width=0.5pt] (1.60,-4.00) rectangle (2.00,-3.60); \fill[latPad,draw=white,line width=0.5pt] (2.00,-4.00) rectangle (2.40,-3.60); \fill[latPad,draw=white,line width=0.5pt] (2.40,-4.00) rectangle (2.80,-3.60); \fill[latPad,draw=white,line width=0.5pt] (2.80,-4.00) rectangle (3.20,-3.60); \fill[latPad,draw=white,line width=0.5pt] (3.20,-4.00) rectangle (3.60,-3.60); \fill[latPad,draw=white,line width=0.5pt] (3.60,-4.00) rectangle (4.00,-3.60); \fill[latPad,draw=white,line width=0.5pt] (4.00,-4.00) rectangle (4.40,-3.60); \fill[latPad,draw=white,line width=0.5pt] (4.40,-4.00) rectangle (4.80,-3.60);
		\node[anchor=east,font=\scriptsize,inner sep=1.5pt] at (-0.06,-3.80) {Article ~8};
		\fill[latArt,draw=white,line width=0.5pt] (0.00,-4.40) rectangle (0.40,-4.00); \fill[latPad,draw=white,line width=0.5pt] (0.40,-4.40) rectangle (0.80,-4.00); \fill[latPad,draw=white,line width=0.5pt] (0.80,-4.40) rectangle (1.20,-4.00); \fill[latPad,draw=white,line width=0.5pt] (1.20,-4.40) rectangle (1.60,-4.00); \fill[latPad,draw=white,line width=0.5pt] (1.60,-4.40) rectangle (2.00,-4.00); \fill[latPad,draw=white,line width=0.5pt] (2.00,-4.40) rectangle (2.40,-4.00); \fill[latPad,draw=white,line width=0.5pt] (2.40,-4.40) rectangle (2.80,-4.00); \fill[latPad,draw=white,line width=0.5pt] (2.80,-4.40) rectangle (3.20,-4.00); \fill[latPad,draw=white,line width=0.5pt] (3.20,-4.40) rectangle (3.60,-4.00); \fill[latPad,draw=white,line width=0.5pt] (3.60,-4.40) rectangle (4.00,-4.00); \fill[latPad,draw=white,line width=0.5pt] (4.00,-4.40) rectangle (4.40,-4.00); \fill[latPad,draw=white,line width=0.5pt] (4.40,-4.40) rectangle (4.80,-4.00);
		\node[anchor=east,font=\scriptsize,inner sep=1.5pt] at (-0.06,-4.20) {Article ~9};
		\fill[latAnx,draw=white,line width=0.5pt] (0.00,-4.80) rectangle (0.40,-4.40); \fill[latAnx,draw=white,line width=0.5pt] (0.40,-4.80) rectangle (0.80,-4.40); \fill[latAnx,draw=white,line width=0.5pt] (0.80,-4.80) rectangle (1.20,-4.40); \fill[latPad,draw=white,line width=0.5pt] (1.20,-4.80) rectangle (1.60,-4.40); \fill[latPad,draw=white,line width=0.5pt] (1.60,-4.80) rectangle (2.00,-4.40); \fill[latPad,draw=white,line width=0.5pt] (2.00,-4.80) rectangle (2.40,-4.40); \fill[latPad,draw=white,line width=0.5pt] (2.40,-4.80) rectangle (2.80,-4.40); \fill[latPad,draw=white,line width=0.5pt] (2.80,-4.80) rectangle (3.20,-4.40); \fill[latPad,draw=white,line width=0.5pt] (3.20,-4.80) rectangle (3.60,-4.40); \fill[latPad,draw=white,line width=0.5pt] (3.60,-4.80) rectangle (4.00,-4.40); \fill[latPad,draw=white,line width=0.5pt] (4.00,-4.80) rectangle (4.40,-4.40); \fill[latPad,draw=white,line width=0.5pt] (4.40,-4.80) rectangle (4.80,-4.40);
		\node[anchor=east,font=\scriptsize,inner sep=1.5pt] at (-0.06,-4.60) {Annex~I};
		\fill[latAnx,draw=white,line width=0.5pt] (0.00,-5.20) rectangle (0.40,-4.80); \fill[latAnx,draw=white,line width=0.5pt] (0.40,-5.20) rectangle (0.80,-4.80); \fill[latAnx,draw=white,line width=0.5pt] (0.80,-5.20) rectangle (1.20,-4.80); \fill[latAnx,draw=white,line width=0.5pt] (1.20,-5.20) rectangle (1.60,-4.80); \fill[latPad,draw=white,line width=0.5pt] (1.60,-5.20) rectangle (2.00,-4.80); \fill[latPad,draw=white,line width=0.5pt] (2.00,-5.20) rectangle (2.40,-4.80); \fill[latPad,draw=white,line width=0.5pt] (2.40,-5.20) rectangle (2.80,-4.80); \fill[latPad,draw=white,line width=0.5pt] (2.80,-5.20) rectangle (3.20,-4.80); \fill[latPad,draw=white,line width=0.5pt] (3.20,-5.20) rectangle (3.60,-4.80); \fill[latPad,draw=white,line width=0.5pt] (3.60,-5.20) rectangle (4.00,-4.80); \fill[latPad,draw=white,line width=0.5pt] (4.00,-5.20) rectangle (4.40,-4.80); \fill[latPad,draw=white,line width=0.5pt] (4.40,-5.20) rectangle (4.80,-4.80);
		\node[anchor=east,font=\scriptsize,inner sep=1.5pt] at (-0.06,-5.00) {Annex~II};
		\fill[latAnx,draw=white,line width=0.5pt] (0.00,-5.60) rectangle (0.40,-5.20); \fill[latAnx,draw=white,line width=0.5pt] (0.40,-5.60) rectangle (0.80,-5.20); \fill[latAnx,draw=white,line width=0.5pt] (0.80,-5.60) rectangle (1.20,-5.20); \fill[latPad,draw=white,line width=0.5pt] (1.20,-5.60) rectangle (1.60,-5.20); \fill[latPad,draw=white,line width=0.5pt] (1.60,-5.60) rectangle (2.00,-5.20); \fill[latPad,draw=white,line width=0.5pt] (2.00,-5.60) rectangle (2.40,-5.20); \fill[latPad,draw=white,line width=0.5pt] (2.40,-5.60) rectangle (2.80,-5.20); \fill[latPad,draw=white,line width=0.5pt] (2.80,-5.60) rectangle (3.20,-5.20); \fill[latPad,draw=white,line width=0.5pt] (3.20,-5.60) rectangle (3.60,-5.20); \fill[latPad,draw=white,line width=0.5pt] (3.60,-5.60) rectangle (4.00,-5.20); \fill[latPad,draw=white,line width=0.5pt] (4.00,-5.60) rectangle (4.40,-5.20); \fill[latPad,draw=white,line width=0.5pt] (4.40,-5.60) rectangle (4.80,-5.20);
		\node[anchor=east,font=\scriptsize,inner sep=1.5pt] at (-0.06,-5.40) {Annex~III};
		\node[font=\tiny,inner sep=1pt] at (0.20,-5.76) {1};
		\node[font=\tiny,inner sep=1pt] at (0.60,-5.76) {2};
		\node[font=\tiny,inner sep=1pt] at (1.00,-5.76) {3};
		\node[font=\tiny,inner sep=1pt] at (1.40,-5.76) {4};
		\node[font=\tiny,inner sep=1pt] at (1.80,-5.76) {5};
		\node[font=\tiny,inner sep=1pt] at (2.20,-5.76) {6};
		\node[font=\tiny,inner sep=1pt] at (2.60,-5.76) {7};
		\node[font=\tiny,inner sep=1pt] at (3.00,-5.76) {8};
		\node[font=\tiny,inner sep=1pt] at (3.40,-5.76) {9};
		\node[font=\tiny,inner sep=1pt] at (3.80,-5.76) {10};
		\node[font=\tiny,inner sep=1pt] at (4.20,-5.76) {11};
		\node[font=\tiny,inner sep=1pt] at (4.60,-5.76) {12};
		\node[font=\scriptsize] at (2.40,-6.08) {paragraph position within section $\rightarrow$};
		\node[font=\scriptsize,rotate=90] at (-1.52,-2.80) {section (lattice row)};
	\end{tikzpicture}
	
	\smallskip
	{\scriptsize
		\tikz{\fill[latPre] (0,0) rectangle (0.8em,0.8em);}~Preamble\hspace{1.5ex}
		\tikz{\fill[latRec] (0,0) rectangle (0.8em,0.8em);}~Recitals\hspace{1.5ex}
		\tikz{\fill[latArt] (0,0) rectangle (0.8em,0.8em);}~Article\hspace{1.5ex}
		\tikz{\fill[latAnx] (0,0) rectangle (0.8em,0.8em);}~Annex\hspace{1.5ex}
		\tikz{\fill[latPad,draw=black!15] (0,0) rectangle (0.8em,0.8em);}~Padding (masked)}
	\caption{Semantic-lattice tiling of a EUR-Lex act (32007R0458). Rows index
		sections recovered by the hierarchy parser (Section \ref{subsec:parsing}), columns index paragraph position within a section, and each occupied cell holds the frozen mT5 embedding of its paragraph. Grey cells are masked padding. Only the occupied $14{\times}12$ region of the full $48{\times}32$ lattice is shown.}
	\label{fig:lattice}
	\label{fig:lattice}
\end{figure}
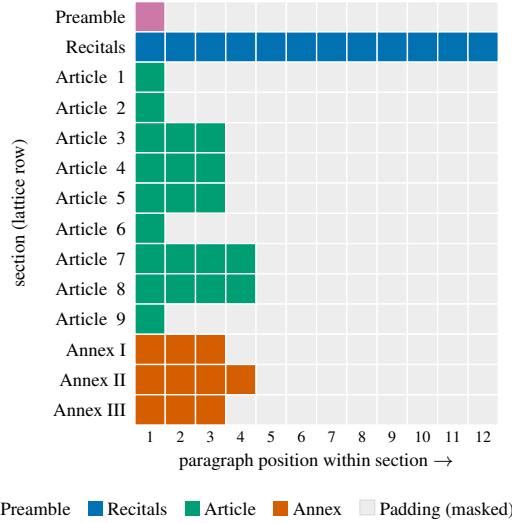

We represent the parsed hierarchy as a fixed two-dimensional grid of size $H \times W = 48 \times 32$, where rows index sections and columns index paragraph positions within a section (Figure~\ref{fig:lattice}). This grid is the substrate of an NCA, a dynamical computational model that learns a single shared update rule through which cells self-organize over a lattice: each cell holds a 768-dimensional state, initialized to its paragraph's frozen embedding, and is updated only from its immediate neighbourhood. We set the grid dimensions from a scan of the training corpus, choosing a grid that accommodates the large majority of documents without inflating the number of empty cells. Because a fixed grid cannot cover every document, we adopt a \textit{clamp-and-merge} policy rather than truncating: a paragraph whose section or position index
falls outside the grid is clamped into the last row or column, and a cell receiving several paragraphs stores their mean vector. \textit{Tiling} scatters each paragraph vector into its cell
and records a binary occupancy mask, and masked cells never contribute to
updates; \textit{un-tiling} broadcasts the updated cell state back to every paragraph mapped to that cell. Both operations are differentiable, so the consolidator trains end to end and every paragraph receives a score. Because the axes are semantic, a cell's neighbors are its adjacent paragraphs within a section and the paragraphs at the same relative position in neighboring
sections.

\subsection{Supervision: a Greedy Extractive Oracle}
\label{sec:oracle}
Following standard practice in extractive summarization
\citep{nallapati2017summarunner, liu-lapata-2019-text}, we construct a greedy oracle per document to serve as the training target. Let $D = \{p_1, \dots, p_N\}$ be the $N$ paragraphs of a document $D$, and $R$ its reference summary. The quality of a selection $S \subseteq D$ against the
reference $R$ is
\begin{equation}
	\label{eq1}
	g(S) = \frac{1}{2}\Bigl(
	\text{R-1}_{F_1}(S, R) + \text{R-2}_{F_1}(S, R)
	\Bigr),
\end{equation}
where $\text{R-1}_{F_1}$ and $\text{R-2}_{F_1}$ denote unigram and bigram \textsc{Rouge} $F_1$ respectively. Starting from an empty selection $S_0 = \varnothing$, we write $\mathcal{R}_t = D \setminus S_{t-1}$ for the paragraphs still available at step $t$, and add the one whose marginal gain in $g$ is largest,
\begin{equation}
	\pi_t = \operatorname*{arg\,max}_{p \, \in \, \mathcal{R}_t}
	\Bigl[\, g\bigl(S_{t-1} \cup \{p\}\bigr) - g(S_{t-1}) \,\Bigr],
\end{equation}
setting $S_t = S_{t-1} \cup \{\pi_t\}$ and terminating when no remaining paragraph yields a positive gain or when $|S_t| = 40$. This cap is a safeguard: the zero-gain criterion terminates first for the large majority of training documents, and where the cap does bind, it removes only the lowest-gain tail of the supervision sequence. The procedure yields both a selection set $S$ and the order in which its members were accepted. It is the ordered sequence $\pi = (\pi_1, \dots, \pi_K)$, with $K = |S| \leq N$, that provides the supervision signal for the warm-start stage (Section \ref{sec:warmstart}).

\subsection{Model: LexLattice}
\label{subsec:lexlattice}
Figure~\ref{fig:model} shows the full pipeline; we describe each component in turn, using the notation of the figure throughout.

\paragraph{Frozen multilingual encoder.}
Every paragraph is embedded with a frozen mT5-base encoder as the attention-masked mean of its final-layer token states (768-d; $\leq$256 tokens per chunk, with longer paragraphs mean-pooling their chunk embeddings), yielding the $N \times 768$ paragraph states that are tiled onto the lattice. Frozen encoder serves three purposes: (i) it holds the shared multilingual representation space fixed as an experimental constant, (ii) it concentrates all trainable capacity in the consolidator and readout, and (iii) it allows embeddings to be computed once and cached.

\paragraph{Masked 2D NCA.}
The consolidator applies the same local update to the lattice state $h^t$ for $T{=}8$ steps (Figure~\ref{fig:model}, right). A perception stage concatenates each cell with a learned depthwise (DW) $3{\times}3$ convolution of its
neighbourhood, $[\,h^t \Vert \mathrm{DWConv}\,h^t\,]$
($768 \to 1{,}536$ channels), and an update MLP of two $1{\times}1$
convolutions ($1{,}536 \to 256 \to 768$, ReLU) produces a residual state
change. The final projection is zero-initialized, so the untrained automata is the identity map. The update is deterministic and is multiplied by the occupancy mask $M$ at every step, so padding cells never contribute to, or drift from, the zero state. Salience evidence thus propagates locally both \textit{along} sections (adjacent
paragraphs) and \textit{across} structurally parallel positions in neighboring sections. Optionally, a learned language embedding ($24 \times 768$) is added to every occupied cell before the first step. The consolidator and readout together hold ${\approx}1.80$M trainable parameters: the frozen encoder does all representation work, and the automata only consolidates.

\paragraph{Capacity-matched 1D control.} \label{sec:model_1dnca}
To isolate the contribution of the second, structural axis we train an otherwise identical 1D-NCA: the same perception and update architecture with width-3 1D convolutions over paragraphs in reading order, and the same step count, hidden width, initialization, readout, loss, and training loop. Any difference between the two systems is attributable to lattice geometry alone.

\paragraph{Readout and selection.}
A residual readout scores each paragraph from the concatenation of its initial
and consolidated states, $[\,h^0 \Vert h^T\,]$:
$\mathrm{Linear}(1{,}536 \to 768) \to \mathrm{GELU} \to
\mathrm{Linear}(768 \to 1)$, yielding a scalar salience score $s_i$. At
inference, paragraphs are ranked by $s_i$ and greedily accepted until the
cumulative word count reaches the reference summary's length; accepted paragraphs are emitted in document order.

\section{Methodology}
\label{sec:methodology}

Stage~1 trains the model that we evaluate as \textsc{LexLattice}; Stage~2 then asks whether reinforcement learning provides any additional value.

\subsection{Stage 1: Listwise Warm-Start}
\label{sec:warmstart}
The readout assigns a score to every paragraph of the document, giving
$s_1, \dots, s_N$. These scores define a Plackett--Luce distribution over paragraphs \citep{luce1959individual, 10.2307/2346567}: a paragraph is drawn without replacement with probability proportional to $\exp(s_i)$, so an ordered selection of $K$ paragraphs has likelihood
\begin{equation}
	\label{eq:plack_luce}
	P(\pi) = \prod_{t=1}^{K}
	\frac{\exp\!\bigl(s_{\pi_t}\bigr)}
	{\sum_{j \in \mathcal{R}_t} \exp\!\bigl(s_j\bigr)},
\end{equation}
with $\mathcal{R}_t = D \setminus S_{t-1}$ as above, so that
$\mathcal{R}_1 = D$ and each factor renormalizes over the paragraphs not yet taken. Taking the oracle sequence of Section~\ref{sec:oracle} as the target, we minimize the Plackett-Luce negative log-likelihood of that ordering (a
listwise ranking loss, equivalent to ListMLE \citep{listmle2008},
\begin{equation}
	\mathcal{L}_{\mathrm{PL}} = -\sum_{t=1}^{K}
	\Bigl[\, s_{\pi_t} - \log \!\!\sum_{j \in \mathcal{R}_t} \exp(s_j) \Bigr],
\end{equation}
which we evaluate with a vectorized reverse log-cumulative-sum-exponential (log-cumsum-exp) rather than a per-step loop. The model therefore scores all $N$ paragraphs while the target constrains the first $K$, leaving the ordering of the unselected remainder free.

Because the listwise objective is invariant to a constant shift of all scores, it fixes only their relative order. We therefore add a binary cross-entropy term $\mathcal{L}_{\mathrm{BCE}}$ over all $N$ paragraphs, with oracle membership as the label, which gives the readout an absolute reference point:
\begin{equation}
	\mathcal{L} = \mathcal{L}_{\mathrm{PL}}
	+ \lambda \, \mathcal{L}_{\mathrm{BCE}},
	\qquad \lambda = 0.3 .
\end{equation}

We set $\lambda = 0.3$ without tuning. The role of $\mathcal{L}_{\mathrm{BCE}}$
is only to resolve the shift-invariance of $\mathcal{L}_{\mathrm{PL}}$ by
anchoring the absolute scale of the scores, which any moderate positive weight
accomplishes; a value well below one keeps the listwise ranking term dominant,
so membership classification refines rather than overrides the learned
ordering. Because every model variant is trained with the same $\lambda$, our
comparisons are unaffected by its precise value.

\subsection{Stage 2: Reinforcement Fine-Tuning}
\label{sec:rl}
Stage~1 optimizes agreement with an oracle sequence, a proxy for the quantity
we actually care about: the quality of the extract the selector finally emits.
Stage~2 optimizes that quantity directly with REINFORCE Leave-One-Out
\citep{kool2019buy, ahmadian-etal-2024-back}, testing whether the proxy leaves
headroom rather than serving as a component the model depends on.

Let $\pi \sim P_\theta$ denote an extract sampled from the Plackett--Luce
policy of Eq.~\ref{eq:plack_luce}: paragraphs are drawn sequentially without replacement until
the word budget is met, then assembled in document order and truncated to the
budget, exactly matching the evaluation-time selector. We maximize the
expected reward of the sampled extract,
\begin{equation}
	\mathcal{J}(\theta) = \mathbb{E}_{\pi \sim P_\theta}\!\bigl[\, r(\pi) \,\bigr],
\end{equation}
where $r(\pi) \in [0,1]$ scores the extract against the reference $R$ as the
mean of three \textsc{Rouge} $F_1$ variants,
\begin{equation}
	r(\pi) = \tfrac{1}{3}\!\!\sum_{v \, \in \, \{1,\,2,\,\mathrm{Lsum}\}}\!\!
	\text{R-}v_{F_1}(\pi, R).
\end{equation}

\paragraph{Language-normalized reward.}
Raw reward magnitudes differ systematically across languages, so before forming
the gradient we standardize $r$ per language $\ell$ using running estimates
$\mu_\ell, \sigma_\ell$ maintained online,
\begin{equation}
	\tilde{r}(\pi) = \frac{r(\pi) - \mu_\ell}{\sigma_\ell},
\end{equation}
which prevents high-scoring languages from dominating the update.

\paragraph{Leave-one-out advantage.}
For each document we draw $G{=}4$ rollouts $\pi_1, \dots, \pi_G$ and use the
mean normalized reward of the remaining $G{-}1$ as each rollout's baseline,
giving a critic-free, unbiased advantage $A_g$ where,
\begin{equation}
	A_g = \tilde{r}(\pi_g) - \frac{1}{G-1}\sum_{g' \neq g} \tilde{r}(\pi_{g'}).
\end{equation}
The resulting policy-gradient estimate is
\begin{equation}
	\nabla_\theta \mathcal{J} \;\approx\;
	\sum_{g=1}^{G} A_g \, \nabla_\theta \log P_\theta(\pi_g),
\end{equation}
where $\log P_\theta(\pi_g)$ is the Plackett-Luce log-likelihood of Eq.~(1),
differentiated through the readout scores.

\section{Experiments}
\label{sec:experiments}

Experiments run on a single workstation with an AMD Ryzen Threadripper 5975WX CPU and three NVIDIA RTX A6000 GPUs (49 GB VRAM each), with all models implemented in PyTorch \citep{pytorch}. Frozen encoder outputs are cached once, after which the warm-start, RL, transfer, and ablation runs proceed without the encoder in the loop. RLOO is the most expensive stage at $\approx$8.4 GPU-hours; warm-start converges within roughly one epoch. All runs use seed 0.

\section{Results and Discussion}
\label{sec:res_disc}

\subsection{Multilingual Analysis}
\label{sec:results}

\begin{table}[!htbp]
	\centering
	\footnotesize
	\setlength{\tabcolsep}{3pt}
	\caption{Per-language results on EUR-Lex-Sum (R1: ROUGE-1, R2: ROUGE-2, RL:
	ROUGE-L, BS: BERTScore-F1). Best per metric per language in \textbf{bold}. LexLattice is our 2D-NCA over the semantic lattice; 1D-NCA is the capacity-matched control (Section~\ref{sec:model_1dnca}).
	$^\dagger$\citet{aumiller-etal-2022-eur}; $^\ddagger$LexT5
	\citep{t-y-s-s-etal-2024-lexsumm}; $^\S$SLM benchmark
	\citep{slms_legislative_sum}.}
	\begin{tabular}{@{}lrrrr@{}}
		\toprule
		Model & R1 & R2 & RL & BS \\
		\midrule
		\multicolumn{5}{@{}l}{\textit{English}}\\
		LexRank$^\dagger$              & 26.68 & 13.64 & 13.58 & -- \\
		SLED-LexT5$^\ddagger$          & 37.40 & 19.30 & 34.30 & 67.60 \\
		Unlim.\,LexT5$^\ddagger$       & 37.90 & 19.10 & 34.10 & 67.50 \\
		LLaMA-3.1-8B-Inst$^\S$         & 47.26 & 21.54 & 21.84 & 72.39 \\
		Qwen3-4B$^\S$                  & 43.07 & 17.59 & 20.45 & 74.14 \\
		1D-NCA (ours)                  & 52.98 & 22.83 & 32.63 & 72.05 \\
		LexLattice (ours)              & 54.81 & \textbf{24.64} & 36.04 & 73.15 \\
		LexLattice\,+\,RLOO (ours)     & \textbf{54.96} & 24.57 & \textbf{36.57} & \textbf{74.28} \\
		\midrule
		\multicolumn{5}{@{}l}{\textit{French}}\\
		LexRank$^\dagger$              & 32.35 & 18.00 & 15.16 & -- \\
		Qwen3-4B$^\S$                  & 51.61 & 27.02 & 25.02 & 79.22 \\
		Qwen2.5-3B-Inst$^\S$           & 51.70 & 26.45 & 25.11 & 78.86 \\
		1D-NCA (ours)                  & 57.68 & 27.21 & 24.98 & 73.27 \\
		LexLattice (ours)              & \textbf{58.75} & 28.55 & 27.23 & 77.72 \\
		LexLattice\,+\,RLOO (ours)     & 58.62 & \textbf{28.67} & \textbf{28.78} & \textbf{80.49} \\
		\midrule
		\multicolumn{5}{@{}l}{\textit{German}}\\
		LexRank$^\dagger$              & 26.72 & 13.75 & 12.56 & -- \\
		Qwen2.5-3B-Inst$_{\text{ext}}^\S$ & 44.58 & 18.86 & 19.20 & 72.35 \\
		Qwen3-0.6B$_{\text{ext}}^\S$   & 45.11 & 18.14 & 17.12 & 70.44 \\
		Qwen3-4B$^\S$                  & 44.72 & 21.40 & 22.79 & \textbf{76.99} \\
		1D-NCA (ours)                  & 48.32 & 20.32 & 19.28 & 71.92 \\
		LexLattice (ours)              & 50.31 & 23.03 & 23.24 & 73.88 \\
		LexLattice\,+\,RLOO (ours)     & \textbf{50.64} & \textbf{23.06} & \textbf{25.14} & 76.76 \\
		\midrule
		\multicolumn{5}{@{}l}{\textit{Italian}}\\
		LexRank$^\dagger$              & 28.57 & 14.24 & 12.90 & -- \\
		Qwen3-0.6B$_{\text{ext}}^\S$   & 46.33 & 18.63 & 17.18 & 72.23 \\
		Qwen2.5-3B$^\S$                & 45.74 & 20.90 & 21.94 & 77.33 \\
		Qwen2.5-1.5B$^\S$              & 42.69 & 19.15 & 21.80 & 77.42 \\
		1D-NCA (ours)                  & 49.70 & 21.49 & 20.61 & 73.04 \\
		LexLattice (ours)              & 51.13 & 22.38 & 22.29 & 77.30 \\
		LexLattice\,+\,RLOO (ours)     & \textbf{51.42} & \textbf{23.29} & \textbf{22.61} & \textbf{78.21} \\
		\midrule
		\multicolumn{5}{@{}l}{\textit{Spanish}}\\
		LexRank$^\dagger$              & 28.34 & 17.12 & 15.23 & -- \\
		Qwen2.5-3B$^\S$                & 53.37 & 28.92 & 28.27 & \textbf{78.92} \\
		Qwen2.5-3B-Inst$^\S$           & 53.08 & 28.50 & 28.83 & 78.83 \\
		1D-NCA (ours)                  & 56.68 & 28.07 & 26.71 & 73.35 \\
		LexLattice (ours)              & 58.41 & 29.44 & 28.48 & 75.87 \\
		LexLattice\,+\,RLOO (ours)     & \textbf{58.44} & \textbf{30.17} & \textbf{29.00} & 76.70 \\
		\midrule
		\multicolumn{5}{@{}l}{\textit{Portuguese}}\\
		LexRank$^\dagger$              & 30.67 & 17.20 & 15.20 & -- \\
		Qwen3-4B-Base$^\S$             & 51.97 & 26.92 & 26.88 & \textbf{78.71} \\
		Qwen2.5-3B-Inst$^\S$           & 51.66 & 27.05 & 27.38 & 78.22 \\
		Qwen3-4B$^\S$                  & 51.16 & 27.08 & 27.21 & 77.92 \\
		1D-NCA (ours)                  & 55.86 & 26.57 & 24.84 & 74.18 \\
		LexLattice (ours)              & 57.52 & 27.68 & 28.13 & 76.38 \\
		LexLattice\,+\,RLOO (ours)     & \textbf{57.55} & \textbf{27.95} & \textbf{28.75} & 78.28 \\
		\bottomrule
	\end{tabular}
	\label{tab:main-results}
\end{table}

Table~\ref{tab:main-results} reports per-language results across six languages,
and three patterns hold consistently. \textit{First}, the structural axis pays
off: LexLattice improves over the 1D-NCA on every metric in every language, by $1.7$-$2.0$ ROUGE-1 and up to $2.7$ ROUGE-2,
confirming that the section$\times$paragraph geometry drives the gain. \textit{Second}, LexLattice+RLOO attains the best ROUGE-1,
ROUGE-2, and ROUGE-L in all six languages, ahead of instruction-tuned baselines
as large as 8B parameters despite carrying only $1.8$M trainable parameters over
a frozen encoder. The margin over the strongest prior system reaches $7.7$
ROUGE-1 in English. \textit{Third}, RLOO adds marginal gains to ROUGE-1 or ROUGE-2 metrics. The
warm-start already captures most of that quality, but yields a consistent
ROUGE-L gain that secures the ROUGE-L lead: in Spanish, for instance, only after
RLOO does our model surpass the strongest baseline on this metric. The BERTScore picture is more balanced: LexLattice is strongest in
English, French, and Italian, while a larger Qwen baseline edges it in German,
Spanish, and Portuguese, in most cases by under a point.

\subsection{Cross-Lingual Transfer}
\label{sec:transfer}
If the consolidator operates on the encoder's shared semantic geometry rather
than on language-specific surface cues, a model trained on one set of languages
should transfer to unseen ones with little loss. We test this in two settings.
In both, the language embedding is omitted, since it is undefined for a language
never seen in training, and we report \textbf{retention}: the ratio of a
transferred model's ROUGE-2 to that of a model jointly supervised on the target
language, where a value of $1$ denotes lossless transfer.

\paragraph{Held-out languages.}
We split the languages at the median training-document count into HIGH (12
languages, 12{,}473 documents) and LOW (12 languages, including Irish with only
16), train the 2D-NCA on HIGH alone, and evaluate zero-shot on LOW.
Table~\ref{tab:transfer} reports the result. Transfer is near-lossless
throughout: retention averages $0.993$ and never falls below $0.974$, and for
five of the twelve languages the zero-shot model matches or exceeds its
jointly-supervised counterpart. A model which has never seen a language can
perform nearly equal to the one trained on it indicates that the consolidation rule the automata learns is largely language-independent.

\begin{table}[t]
	\centering
	\footnotesize
	\setlength{\tabcolsep}{5pt}
	\caption{Zero-shot transfer on the 12 held-out (LOW) languages, ROUGE-2.
	\textit{Transfer}: 2D-NCA trained on HIGH only; \textit{Joint}: trained on all
	languages. \textit{Retention} = Transfer\,/\,Joint; ${\geq}1$ indicates no loss. Sorted by retention.}
	\begin{tabular}{@{}lrrr@{}}
		\toprule
		Language & Transfer & Joint & Retention \\
		\midrule
		Polish      & 21.89 & 21.64 & 1.012 \\
		Irish       & 25.61 & 25.48 & 1.005 \\
		Hungarian   & 30.41 & 30.34 & 1.002 \\
		Lithuanian  & 20.10 & 20.07 & 1.002 \\
		Latvian     & 24.03 & 24.00 & 1.001 \\
		Maltese     & 24.81 & 24.92 & 0.996 \\
		Bulgarian   & 16.95 & 17.06 & 0.993 \\
		Romanian    & 21.97 & 22.22 & 0.989 \\
		Estonian    & 17.89 & 18.11 & 0.988 \\
		Slovak      & 23.32 & 23.85 & 0.978 \\
		Croatian    & 19.83 & 20.32 & 0.976 \\
		Slovenian   & 19.21 & 19.73 & 0.974 \\
		\midrule
		Mean        & 22.17 & 22.31 & 0.993 \\
		\bottomrule
	\end{tabular}
	\label{tab:transfer}
\end{table}

\paragraph{English${\rightarrow}$Spanish.}
As a stricter case mirroring the cross-lingual baseline of the original
EUR-Lex-Sum study, we train LexLattice on English alone, select on English
validation, and evaluate zero-shot on Spanish, so the target language is never
seen before test time (Table~\ref{tab:crosslingual}). Among methods with access
only to the English source, zero-shot LexLattice improves over both the
translate-then-summarize LED pipeline and English LexRank by a wide margin on
every metric. It also comes within $1.8$ ROUGE-1 of the Spanish-supervised
model, reinforcing the finding above: transferring from English costs little
even when no target-language signal is available at any stage.

\begin{table}[t]
	\centering
	\footnotesize
	\setlength{\tabcolsep}{4pt}
	\caption{Cross-lingual English${\rightarrow}$Spanish results. Upper block:
	English-source-only methods (\textbf{bold} = best). Lower block: Spanish-side
	references (italic). $^\dagger$\citet{aumiller-etal-2022-eur}.}
	\begin{tabular}{@{}lrrrr@{}}
		\toprule
		Model & R1 & R2 & RL & BS \\
		\midrule
		LED$^\dagger$              & 31.14 & 13.01 & 16.20 & -- \\
		LexRank-EN$^\dagger$       & 39.44 & 20.02 & 18.73 & -- \\
		LexLattice-EN (zero-shot)  & \textbf{56.63} & \textbf{28.33} & \textbf{27.68} & \textbf{73.63} \\
		\midrule
		\textit{LexRank-ES}$^\dagger$  & \textit{28.34} & \textit{17.12} & \textit{15.23} & \textit{--} \\
		\textit{Oracle-Abs}$^\dagger$  & \textit{54.55} & \textit{41.01} & \textit{45.06} & \textit{--} \\
		\textit{LexLattice-ES} (sup.)  & \textit{58.41} & \textit{29.44} & \textit{31.24} & \textit{74.87} \\
		\bottomrule
	\end{tabular}
	\label{tab:crosslingual}
\end{table}

\subsection{LLM-as-a-Judge Evaluation}
\label{sec:lemaj}
\begin{table}[t]
	\centering
	\small
	\setlength{\tabcolsep}{5pt}
	\caption{LLM-as-a-judge evaluation on the full test set (all 24 languages,
	candidates blinded), macro-averaged over languages. \emph{Coverage}: fraction
	of reference LDPs recovered; \emph{relevance}: fraction of candidate content
	judged pertinent; LDP-$F_1$: their harmonic mean. $^{\dagger}$/$^{\ddagger}$:
	improvement over 1D-NCA / LexLattice (paired Wilcoxon, per-document,
	$p<0.05$).}
	\begin{tabular}{@{}lccc@{}}
		\toprule
		Model & Coverage & Relevance & LDP-$F_1$ \\
		\midrule
		1D-NCA              & 37.3 & 68.0 & 45.1 \\
		LexLattice          & 44.4$^{\dagger}$ & 75.3$^{\dagger}$ & 52.9$^{\dagger}$ \\
		LexLattice\,+\,RLOO & \textbf{45.3}$^{\dagger}$ & \textbf{77.2}$^{\dagger\ddagger}$ & \textbf{54.2}$^{\dagger}$ \\
		\bottomrule
	\end{tabular}
	\label{tab:lemaj}
\end{table}
ROUGE under-credits heavily inflected languages and behaves inconsistently
across scripts. We therefore complement it with a reference-based adaptation of
LeMAJ \citep{enguehard-etal-2025-lemaj}, which decomposes each reference into
\textit{Legal Data Points} (LDPs): atomic, self-contained statements of legal
information. A single fixed judge (Claude Sonnet~4.5,
\citealp{anthropic2025sonnet45}) scores each candidate pointwise and blind, with
\textit{coverage} marking every reference LDP as covered, partial, or missing,
and \textit{relevance} marking every candidate unit as relevant, marginal, or
irrelevant (partial credit $0.5$); LDP-$F_1$ is their harmonic mean. Because the
systems are extractive and therefore near-perfectly grounded in the source, the
protocol isolates \emph{selection} quality rather than factual consistency. The judge-based results in Table~\ref{tab:lemaj} demonstrate the ROUGE findings
under an evaluation largely insensitive to inflection and script. The structural
axis accounts for most of the improvement: LexLattice improves LDP-$F_1$ over the
1D control by $7.8$ points ($+17\%$), the majority of it attributable to
coverage ($+7.1$), which suggests that structural adjacency helps the model
recover salient content that a flat pass over paragraphs overlooks. RLOO
contributes a further $1.3$ points, concentrated in relevance, and all
improvements are significant under a per-document paired Wilcoxon test.

\subsection{Coverage Across the Remaining Languages}
\label{sec:rest-languages}
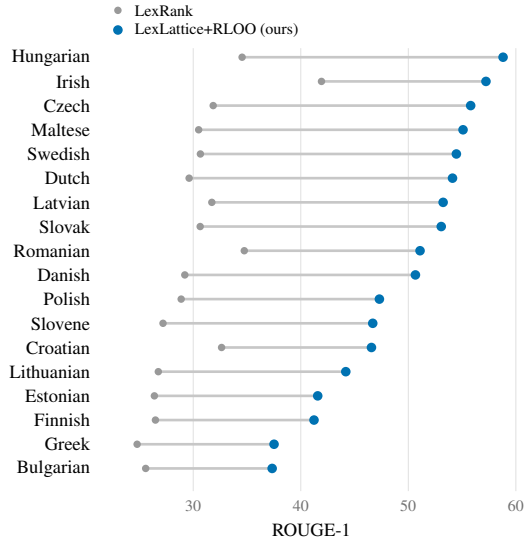
\begin{figure}[!ht]
	\centering
	\begin{tikzpicture}[x=1cm,y=1cm]
		\draw[latG!30,line width=0.3pt] (1.137,0.12) -- (1.137,-5.74);
		\node[font=\tiny,text=black!55,anchor=north] at (1.137,-5.74) {30};
		\draw[latG!30,line width=0.3pt] (2.558,0.12) -- (2.558,-5.74);
		\node[font=\tiny,text=black!55,anchor=north] at (2.558,-5.74) {40};
		\draw[latG!30,line width=0.3pt] (3.979,0.12) -- (3.979,-5.74);
		\node[font=\tiny,text=black!55,anchor=north] at (3.979,-5.74) {50};
		\draw[latG!30,line width=0.3pt] (5.4,0.12) -- (5.4,-5.74);
		\node[font=\tiny,text=black!55,anchor=north] at (5.4,-5.74) {60};
		\node[font=\scriptsize,anchor=north] at (2.7,-6.04) {ROUGE-1};
		\draw[latG!55,line width=1.0pt] (1.783,0.0) -- (5.231,0.0);
		\fill[latG] (1.783,0.0) circle (1.4pt);
		\fill[latB] (5.231,0.0) circle (1.8pt);
		\node[font=\scriptsize,anchor=east] at (-0.10,0.0) {Hungarian};
		\draw[latG!55,line width=1.0pt] (2.832,-0.32) -- (5.005,-0.32);
		\fill[latG] (2.832,-0.32) circle (1.4pt);
		\fill[latB] (5.005,-0.32) circle (1.8pt);
		\node[font=\scriptsize,anchor=east] at (-0.10,-0.32) {Irish};
		\draw[latG!55,line width=1.0pt] (1.401,-0.64) -- (4.803,-0.64);
		\fill[latG] (1.401,-0.64) circle (1.4pt);
		\fill[latB] (4.803,-0.64) circle (1.8pt);
		\node[font=\scriptsize,anchor=east] at (-0.10,-0.64) {Czech};
		\draw[latG!55,line width=1.0pt] (1.209,-0.96) -- (4.702,-0.96);
		\fill[latG] (1.209,-0.96) circle (1.4pt);
		\fill[latB] (4.702,-0.96) circle (1.8pt);
		\node[font=\scriptsize,anchor=east] at (-0.10,-0.96) {Maltese};
		\draw[latG!55,line width=1.0pt] (1.232,-1.28) -- (4.614,-1.28);
		\fill[latG] (1.232,-1.28) circle (1.4pt);
		\fill[latB] (4.614,-1.28) circle (1.8pt);
		\node[font=\scriptsize,anchor=east] at (-0.10,-1.28) {Swedish};
		\draw[latG!55,line width=1.0pt] (1.083,-1.6) -- (4.563,-1.6);
		\fill[latG] (1.083,-1.6) circle (1.4pt);
		\fill[latB] (4.563,-1.6) circle (1.8pt);
		\node[font=\scriptsize,anchor=east] at (-0.10,-1.6) {Dutch};
		\draw[latG!55,line width=1.0pt] (1.383,-1.92) -- (4.439,-1.92);
		\fill[latG] (1.383,-1.92) circle (1.4pt);
		\fill[latB] (4.439,-1.92) circle (1.8pt);
		\node[font=\scriptsize,anchor=east] at (-0.10,-1.92) {Latvian};
		\draw[latG!55,line width=1.0pt] (1.229,-2.24) -- (4.414,-2.24);
		\fill[latG] (1.229,-2.24) circle (1.4pt);
		\fill[latB] (4.414,-2.24) circle (1.8pt);
		\node[font=\scriptsize,anchor=east] at (-0.10,-2.24) {Slovak};
		\draw[latG!55,line width=1.0pt] (1.812,-2.56) -- (4.134,-2.56);
		\fill[latG] (1.812,-2.56) circle (1.4pt);
		\fill[latB] (4.134,-2.56) circle (1.8pt);
		\node[font=\scriptsize,anchor=east] at (-0.10,-2.56) {Romanian};
		\draw[latG!55,line width=1.0pt] (1.026,-2.88) -- (4.073,-2.88);
		\fill[latG] (1.026,-2.88) circle (1.4pt);
		\fill[latB] (4.073,-2.88) circle (1.8pt);
		\node[font=\scriptsize,anchor=east] at (-0.10,-2.88) {Danish};
		\draw[latG!55,line width=1.0pt] (0.978,-3.2) -- (3.597,-3.2);
		\fill[latG] (0.978,-3.2) circle (1.4pt);
		\fill[latB] (3.597,-3.2) circle (1.8pt);
		\node[font=\scriptsize,anchor=east] at (-0.10,-3.2) {Polish};
		\draw[latG!55,line width=1.0pt] (0.738,-3.52) -- (3.509,-3.52);
		\fill[latG] (0.738,-3.52) circle (1.4pt);
		\fill[latB] (3.509,-3.52) circle (1.8pt);
		\node[font=\scriptsize,anchor=east] at (-0.10,-3.52) {Slovene};
		\draw[latG!55,line width=1.0pt] (1.512,-3.84) -- (3.492,-3.84);
		\fill[latG] (1.512,-3.84) circle (1.4pt);
		\fill[latB] (3.492,-3.84) circle (1.8pt);
		\node[font=\scriptsize,anchor=east] at (-0.10,-3.84) {Croatian};
		\draw[latG!55,line width=1.0pt] (0.676,-4.16) -- (3.153,-4.16);
		\fill[latG] (0.676,-4.16) circle (1.4pt);
		\fill[latB] (3.153,-4.16) circle (1.8pt);
		\node[font=\scriptsize,anchor=east] at (-0.10,-4.16) {Lithuanian};
		\draw[latG!55,line width=1.0pt] (0.624,-4.48) -- (2.782,-4.48);
		\fill[latG] (0.624,-4.48) circle (1.4pt);
		\fill[latB] (2.782,-4.48) circle (1.8pt);
		\node[font=\scriptsize,anchor=east] at (-0.10,-4.48) {Estonian};
		\draw[latG!55,line width=1.0pt] (0.638,-4.8) -- (2.733,-4.8);
		\fill[latG] (0.638,-4.8) circle (1.4pt);
		\fill[latB] (2.733,-4.8) circle (1.8pt);
		\node[font=\scriptsize,anchor=east] at (-0.10,-4.8) {Finnish};
		\draw[latG!55,line width=1.0pt] (0.396,-5.12) -- (2.205,-5.12);
		\fill[latG] (0.396,-5.12) circle (1.4pt);
		\fill[latB] (2.205,-5.12) circle (1.8pt);
		\node[font=\scriptsize,anchor=east] at (-0.10,-5.12) {Greek};
		\draw[latG!55,line width=1.0pt] (0.509,-5.44) -- (2.18,-5.44);
		\fill[latG] (0.509,-5.44) circle (1.4pt);
		\fill[latB] (2.18,-5.44) circle (1.8pt);
		\node[font=\scriptsize,anchor=east] at (-0.10,-5.44) {Bulgarian};
		\fill[latG] (0.142,0.62) circle (1.4pt); \node[font=\tiny,anchor=west] at (0.242,0.62) {LexRank};
		\fill[latB] (0.142,0.36) circle (1.8pt); \node[font=\tiny,anchor=west] at (0.242,0.36) {LexLattice+RLOO (ours)};
	\end{tikzpicture}
	\caption{ROUGE-1 on the remaining 18 EUR-Lex-Sum languages (jointly trained,
		sorted by our score). LexRank is the only external baseline available for these
		languages; our final model improves on it by $19.0$ points on average
		($11.8$--$24.6$), in every language. Full per-system, per-metric results are in
		Appendix~\ref{app:full-results}.}
	\label{fig:rest-languages}
\end{figure}

The six languages of Section~\ref{sec:results} are the only ones with
instruction-tuned LLM or SLM variants; for the remaining 18, the zero-shot
multilingual LexRank of \citet{aumiller-etal-2022-eur} is the sole published
baseline. Figure~\ref{fig:rest-languages} shows ROUGE-1 for these languages,
and the gains are uniform: LexLattice model with RLOO improves on LexRank in every one, by $19.0$ points on average and by at least $11.8$, spanning scripts from Latin to Cyrillic (Bulgarian) and Greek. The 1D$\rightarrow$2D and RLOO trends observed
on the high-resource languages persist here as well, with the semantic lattice
providing the bulk of the improvement over the 1D control; full per-system,
per-metric numbers appear in Section~\ref{app:all_results}.

\subsection{Complete Results Across All Languages}
\label{app:all_results}
\noindent Table~\ref{app:full-results} reports the complete per-system,
per-metric scores for the 18 EUR-Lex-Sum languages not shown in
Table~\ref{tab:main-results}, jointly trained on all 24 languages. LexLattice or
its RLOO variant attains the best ROUGE-1 and ROUGE-2 in every language, with the
1D control competitive only in Bulgarian.

\begin{table}[H]
	\centering
	\footnotesize
	\setlength{\tabcolsep}{2.5pt}
	\caption{Full per-language, per-system results on the 18 EUR-Lex-Sum
	languages not shown in Table~\ref{tab:main-results} (jointly trained on all 24
	languages). Best value per metric per language in \textbf{bold}. LexLattice is
	our 2D-NCA; \texttt{+RLOO} adds the reinforcement stage; 1D-NCA is the
	capacity-matched control.}
	\begin{minipage}[t]{0.32\textwidth}\centering
		\begin{tabular}{@{}lrrrr@{}}
			\toprule
			Model & R1 & R2 & RL & BS \\
			\midrule
			\multicolumn{5}{@{}l}{\textit{Bulgarian}}\\
			LexRank & 25.58 & 8.40 & 16.13 & -- \\
			1D-NCA & 36.92 & \textbf{18.06} & \textbf{21.51} & 73.45 \\
			LexLattice & \textbf{37.44} & 17.25 & 21.06 & \textbf{73.54} \\
			\quad+RLOO & 37.34 & 16.17 & 20.26 & 73.30 \\
			\midrule
			\multicolumn{5}{@{}l}{\textit{Croatian}}\\
			LexRank & 32.64 & 12.76 & 13.29 & -- \\
			1D-NCA & 44.58 & 18.58 & 18.09 & 73.40 \\
			LexLattice & 46.32 & 19.85 & 18.34 & 74.11 \\
			\quad+RLOO & \textbf{46.57} & \textbf{20.47} & \textbf{18.42} & \textbf{74.93} \\
			\midrule
			\multicolumn{5}{@{}l}{\textit{Czech}}\\
			LexRank & 31.86 & 16.76 & 14.32 & -- \\
			1D-NCA & 53.10 & 23.88 & 21.10 & 72.79 \\
			LexLattice & 55.49 & \textbf{25.87} & 21.91 & 73.98 \\
			\quad+RLOO & \textbf{55.80} & 25.34 & \textbf{21.96} & \textbf{74.12} \\
			\midrule
			\multicolumn{5}{@{}l}{\textit{Danish}}\\
			LexRank & 29.22 & 13.86 & 13.19 & -- \\
			1D-NCA & 48.84 & 20.15 & 18.90 & 72.64 \\
			LexLattice & 50.31 & 21.07 & 19.55 & \textbf{72.97} \\
			\quad+RLOO & \textbf{50.66} & \textbf{21.30} & \textbf{20.42} & 72.72 \\
			\midrule
			\multicolumn{5}{@{}l}{\textit{Dutch}}\\
			LexRank & 29.62 & 14.76 & 14.73 & -- \\
			1D-NCA & 51.31 & 22.96 & 22.81 & 73.88 \\
			LexLattice & 53.83 & \textbf{25.24} & 23.19 & \textbf{74.33} \\
			\quad+RLOO & \textbf{54.11} & 24.68 & \textbf{23.57} & 73.94 \\
			\midrule
			\multicolumn{5}{@{}l}{\textit{Estonian}}\\
			LexRank & 26.39 & 11.41 & 11.66 & -- \\
			1D-NCA & 40.24 & 16.61 & 16.57 & 72.93 \\
			LexLattice & \textbf{41.58} & 17.97 & 16.88 & 73.10 \\
			\quad+RLOO & \textbf{41.58} & \textbf{17.99} & \textbf{17.16} & \textbf{73.84} \\
			\bottomrule
		\end{tabular}
	\end{minipage}\hfill
	\begin{minipage}[t]{0.32\textwidth}\centering
		\begin{tabular}{@{}lrrrr@{}}
			\toprule
			Model & R1 & R2 & RL & BS \\
			\midrule
			\multicolumn{5}{@{}l}{\textit{Finnish}}\\
			LexRank & 26.49 & 11.68 & 11.80 & -- \\
			1D-NCA & 39.86 & 17.61 & 17.61 & 72.82 \\
			LexLattice & 40.72 & 18.01 & 18.74 & 74.06 \\
			\quad+RLOO & \textbf{41.23} & \textbf{18.64} & \textbf{19.11} & \textbf{75.66} \\
			\midrule
			\multicolumn{5}{@{}l}{\textit{Greek}}\\
			LexRank & 24.79 & 9.45 & 15.46 & -- \\
			1D-NCA & 37.00 & 17.60 & 21.35 & 75.35 \\
			LexLattice & \textbf{38.35} & 17.71 & \textbf{22.03} & 75.55 \\
			\quad+RLOO & 37.52 & \textbf{18.36} & 21.82 & \textbf{75.78} \\
			\midrule
			\multicolumn{5}{@{}l}{\textit{Hungarian}}\\
			LexRank & 34.55 & 19.69 & 15.64 & -- \\
			1D-NCA & 56.50 & 28.20 & 22.13 & 74.20 \\
			LexLattice & \textbf{58.82} & \textbf{30.73} & 22.80 & 75.72 \\
			\quad+RLOO & 58.81 & 30.40 & \textbf{23.14} & \textbf{76.35} \\
			\midrule
			\multicolumn{5}{@{}l}{\textit{Irish}}\\
			LexRank & 41.93 & 17.16 & 15.25 & -- \\
			1D-NCA & 54.94 & 23.31 & 18.99 & 75.64 \\
			LexLattice & 57.11 & 25.12 & 19.04 & 75.67 \\
			\quad+RLOO & \textbf{57.22} & \textbf{25.33} & \textbf{19.47} & \textbf{76.66} \\
			\midrule
			\multicolumn{5}{@{}l}{\textit{Latvian}}\\
			LexRank & 31.73 & 15.77 & 13.15 & -- \\
			1D-NCA & 51.75 & 22.84 & 19.01 & 74.26 \\
			LexLattice & \textbf{53.25} & \textbf{23.94} & 20.70 & 75.55 \\
			\quad+RLOO & 53.24 & 23.74 & \textbf{21.09} & \textbf{76.23} \\
			\midrule
			\multicolumn{5}{@{}l}{\textit{Lithuanian}}\\
			LexRank & 26.76 & 12.45 & 11.59 & -- \\
			1D-NCA & 42.65 & 18.72 & 17.09 & 73.28 \\
			LexLattice & 43.55 & 19.84 & 17.48 & 73.61 \\
			\quad+RLOO & \textbf{44.19} & \textbf{19.92} & \textbf{17.69} & \textbf{74.31} \\
			\bottomrule
		\end{tabular}
	\end{minipage}\hfill
	\begin{minipage}[t]{0.32\textwidth}\centering
		\begin{tabular}{@{}lrrrr@{}}
			\toprule
			Model & R1 & R2 & RL & BS \\
			\midrule
			\multicolumn{5}{@{}l}{\textit{Maltese}}\\
			LexRank & 30.51 & 14.62 & 12.86 & -- \\
			1D-NCA & 53.63 & 23.97 & 19.74 & 75.49 \\
			LexLattice & 54.82 & 24.74 & 20.26 & 76.50 \\
			\quad+RLOO & \textbf{55.09} & \textbf{25.15} & \textbf{20.36} & \textbf{77.47} \\
			\midrule
			\multicolumn{5}{@{}l}{\textit{Polish}}\\
			LexRank & 28.88 & 14.42 & 12.73 & -- \\
			1D-NCA & 46.31 & 21.31 & 18.65 & 72.74 \\
			LexLattice & 47.04 & 21.53 & 19.88 & 73.05 \\
			\quad+RLOO & \textbf{47.31} & \textbf{21.58} & \textbf{21.32} & \textbf{74.66} \\
			\midrule
			\multicolumn{5}{@{}l}{\textit{Romanian}}\\
			LexRank & 34.75 & 15.16 & 14.59 & -- \\
			1D-NCA & 49.93 & 21.02 & 18.59 & 72.96 \\
			LexLattice & 50.97 & \textbf{22.72} & 19.53 & 73.02 \\
			\quad+RLOO & \textbf{51.09} & 22.66 & \textbf{20.17} & \textbf{73.16} \\
			\midrule
			\multicolumn{5}{@{}l}{\textit{Slovak}}\\
			LexRank & 30.65 & 14.94 & 13.14 & -- \\
			1D-NCA & 50.09 & 21.65 & 19.06 & 72.07 \\
			LexLattice & 52.99 & 23.52 & 19.64 & 73.02 \\
			\quad+RLOO & \textbf{53.06} & \textbf{24.15} & \textbf{20.81} & \textbf{74.95} \\
			\midrule
			\multicolumn{5}{@{}l}{\textit{Slovene}}\\
			LexRank & 27.19 & 12.34 & 11.79 & -- \\
			1D-NCA & 43.44 & 18.40 & 17.62 & 72.99 \\
			LexLattice & 45.92 & \textbf{20.65} & 18.11 & 73.62 \\
			\quad+RLOO & \textbf{46.69} & 19.35 & \textbf{20.52} & \textbf{75.40} \\
			\midrule
			\multicolumn{5}{@{}l}{\textit{Swedish}}\\
			LexRank & 30.67 & 15.47 & 14.35 & -- \\
			1D-NCA & 53.23 & 23.42 & 20.73 & 74.92 \\
			LexLattice & 54.45 & \textbf{25.18} & \textbf{22.11} & 75.05 \\
			\quad+RLOO & \textbf{54.47} & 24.82 & 21.52 & \textbf{75.85} \\
			\bottomrule
		\end{tabular}
	\end{minipage}
	\label{app:full-results}
\end{table}

\subsection{Visualizing NCA Salience Evolution}
\label{app:nca-salience}
\begin{figure}[H]
	\centering
	\includegraphics[width=\textwidth]{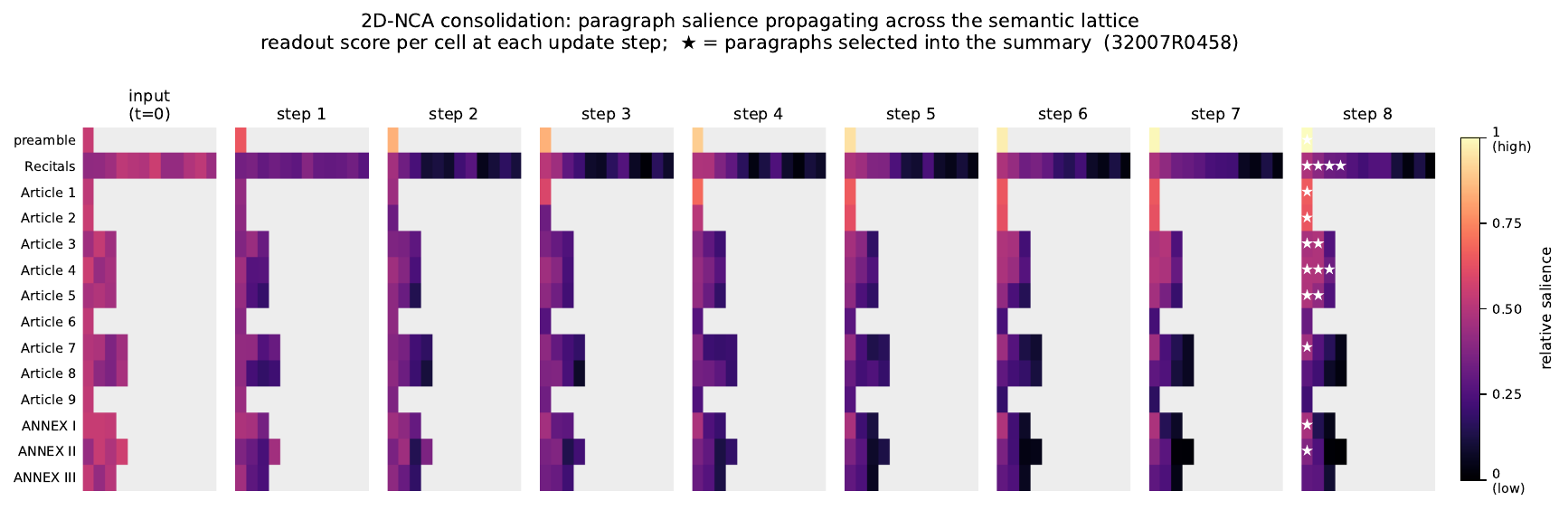}
	\caption{Per-cell readout salience on the semantic lattice at $t{=}0$ and after each of the 8 NCA update steps (single sequential colour ramp; grey = padding). $\star$ marks paragraphs selected into the final extract.}
	\label{fig:nca-salience}
\end{figure}

\noindent Figure~\ref{fig:nca-salience} illustrates the consolidation dynamics of LexLattice on a representative English document. Each panel shows the semantic
lattice (rows: preamble, recitals, articles, annexes; columns: paragraph position within the section; grey cells are padding), coloured by the readout salience of each paragraph, min--max normalized over the whole trajectory. The leftmost panel is the encoder input ($t{=}0$); subsequent panels show the state after each of the $T{=}8$ local NCA update steps. Salience is diffuse at $t{=}0$
and progressively sharpens as information propagates through the $3{\times}3$ neighbourhood, concentrating on a sparse set of cells. Stars in the final panel mark the paragraphs ranked into the summary under the reference-length word budget; these cluster in contiguous lattice regions rather than isolated cells, evidence that the learned dynamics exploit the document's hierarchical structure rather than scoring paragraphs independently.

\section{Conclusion and Future Work}
We introduced LexLattice, an extractive summarizer that casts a legal act's hierarchy as a 2D semantic lattice and consolidates over it with a
masked 2D NCA before selection. Concentrating all trainable capacity in a $1.8$M-parameter consolidator over a frozen multilingual encoder, it attains
state-of-the-art ROUGE scores across all 24 languages of EUR-Lex-Sum, surpassing far larger instruction-tuned baselines, with a uniform per-language evaluation and a broad cross-lingual transfer study. The learned dynamics sharpen diffuse salience onto contiguous, structurally coherent regions of the lattice (Section ~\ref{app:nca-salience}), indicating that explicit consolidation over
document structure is a compact substitute for scale. Proposed extensions of our work include:
applying the lattice to other long, hierarchically organized corpora, both legal
\citep{NEURIPS2022_552ef803} and beyond \citep{kryscinski-etal-2022-booksum}; and
coupling the traceable extract with an abstractive generator to add fluency while
preserving auditability.


\section{Limitations}
LexLattice relies on a rule-based parser to recover document hierarchy, so it presupposes structurally marked text and would degrade on documents without explicit sectioning. Being extractive, it is bounded by the source and a fixed length budget, which caps coverage and forgoes the fluency of abstraction. Finally, for most of the languages the only published baseline is extractive LexRank, so our comparisons rest on limited external baselines.

\section*{Ethical Statement}

This work uses EUR-Lex-Sum, a publicly released corpus of EU legal acts and their official summaries; it contains no personal or private data, and
we use it in accordance with its license. Our aim is to broaden access to law
across the official EU languages, which is of particular benefit to speakers of
lower-resourced languages underserved by English-centric systems. We caution,
however, that automatic summaries of legal text are not legal advice and must
not substitute for the authoritative instruments they condense. Although our
extractive design keeps summaries traceable to the source and limits fabricated
content, it guarantees neither completeness nor faithfulness, and quality is
weaker for the most resource-scarce languages, where reliance should be
correspondingly cautious. Because the method trains only a small consolidator
over a frozen encoder, it also carries a modest computational and energy cost
relative to large generative summarizers.

\section*{Acknowledgments}
This research is supported by NSERC Discovery grant \#194376.

\bibliographystyle{unsrtnat}
\bibliography{references} 

\end{document}